\documentclass[sigconf
]{acmart}

\usepackage{booktabs} 
\usepackage{balance}

\usepackage[ruled]{algorithm2e} 

\SetAlFnt{\small}
\SetAlCapFnt{\small}
\SetAlCapNameFnt{\small}
\SetAlCapHSkip{0pt}
\IncMargin{-\parindent}

\renewcommand\footnotetextcopyrightpermission[1]{} 
\begin{document}
\title{Deep Neural Networks for Bot Detection}

\author{Sneha Kudugunta}
\affiliation{%
  \institution{Indian Institute of Technology, Hyderabad}
  \city{Hyderabad}
  \state{India}
}
\email{cs14btech11020@iith.ac.in}

\author{Emilio Ferrara}
\affiliation{%
  \institution{USC Information Sciences Institute}
  \city{Marina Del Rey} 
  \state{CA, USA} 
}
\email{emiliofe@usc.edu}

\begin{abstract}
The problem of detecting bots, automated social media accounts governed by software but disguising as human users, has strong implications. For example, bots have been used to sway political elections by distorting online discourse, to manipulate the stock market, or to push anti-vaccine conspiracy theories that caused health epidemics. Most techniques proposed to date detect bots at the account level, by processing large amount of social media posts, and leveraging information from network structure, temporal dynamics, sentiment analysis, etc.
In this paper, we propose a deep neural network based on contextual long short-term memory (LSTM) architecture that exploits both content and metadata to detect bots at the tweet level: contextual features are extracted from user metadata and fed as auxiliary input to LSTM deep nets processing the tweet text. 
Another contribution that we make is proposing a technique based on synthetic minority oversampling to generate a large labeled dataset, suitable for deep nets training, from a minimal amount of labeled data (roughly 3,000 examples of sophisticated Twitter bots).
We demonstrate that, from just one single tweet, our architecture can achieve high classification accuracy (AUC $>$ 96\%) in separating bots from humans. 
We apply the same architecture to account-level bot detection, achieving nearly perfect classification accuracy (AUC $>$ 99\%). Our system outperforms previous state of the art while leveraging a small and interpretable set of features yet requiring minimal training data.

\end{abstract}

%

\ccsdesc[500]{Networks~Social media networks}
\ccsdesc[500]{Information systems~Web and social media search}
\ccsdesc[500]{Computer systems organization~Neural networks}





\maketitle

\renewcommand{\shortauthors}{Sneha Kudugunta, Emilio Ferrara}


\section{Introduction}

During the past decade, social media like Twitter and Facebook emerged as a widespread tool for massive-scale and real-time communication. These platforms have been promptly praised by some researchers for their power to democratize discussions~\cite{loader2011networking}, for example by allowing citizens of countries with oppressing regimes to openly discuss social and political issues.
However, due to many recent reports of social media manipulation, including political propaganda, extremism, disinformation, etc., concerns about their abuse are mounting~\cite{gayo2017social}. 

One example of social media manipulation is the use of bots (a.k.a. social bots, or \textit{sybil accounts}), user accounts controlled by software algorithms rather than human users.
Bots have been extensively used for disingenuous purposes, ranging from swaying political opinions to perpetuating scams. Existing social media bots vary in sophistication. Some bots are very simple and merely retweet posts of interest, whereas others are more sophisticated and have the capability to even interact with human users. 

The challenge of bot detection has thus been faced by our research community. To detect these types of social media bots, different approaches have been proposed. Supervised learning in particular exhibited promising results: examples of activity of human users and bots, labeled as such, can be fed to machine learning algorithms; trained models are then used to classify unforeseen accounts, leveraging data obtained e.g., by using the Twitter API: this may help determine the nature of suspicious accounts. Alternatives based on unsupervised learning aimed at identify large-scale behavioral anomalies and associate them to bot accounts.

However, most, if not all, of the successful methods introduced so far detect bots at the account level. This means that, given a record of activity (e.g., a few hundred tweets posted by a user), the algorithm would determine whether the scrutinized account is likely a bot or not. These approaches tend to focus on the overall activity of the account, e.g., the content and sentiment of user posts, the network structure, and the temporal activity patterns. 

Though reasonably successful, account-level bot detection approaches are expensive as they require significant amounts of data from each user to scrutinize, as well as large labeled datasets for training purposes. In contrast with that, most available labeled datasets have at most a few hundreds examples of tweets posted by a few thousands bots. For a comprehensive survey of bot detection methods, we direct the user to ~\cite{ferrara2016rise}.

\subsection{Research questions}
These fundamental limitations pose two research questions, that we try to address in this paper:

\begin{itemize}
\item[RQ1:] Is it possible to accurately predict whether a given tweet has been posted by a bot or human account?
\item[RQ2:] Is it possible to enhance existing labeled datasets to produce more examples of bot and human accounts without the additional (and very expensive) data collection and annotation steps?
\end{itemize}

\subsection{Contributions of this work}
The contributions we provide here aim to address these challenges:

\begin{enumerate}
\item We advance the problem of classifying individual social media accounts from single observations, i.e., determining whether a single tweet comes from a Twitter bot or from a human user. We demonstrate that tweet-level bot detection is possible and can be very accurate: by exploiting both textual features and tweet metadata, we detect bots from single tweet and even exceed the performance of earlier works that make use of a given user's entire profile and recent posting history.

\item As a technical contribution, we introduce the concept of a \textit{Contextual LSTM} (long short-term memory) deep neural network~\cite{hochreiter1997long,gers1999learning}, an architecture that takes both the tweet text and the tweet metadata as an input. Related architectures that use side-information to enhance recurrent model representations have been alluded to by some authors working primarily in language models~\cite{hoang2016incorporating,auli2013joint,mikolov2012context}, but has never been used in the context of social media classification to the best of our knowledge. The proposed architecture allows us to reach state-of-the-art performance in bot detection (over 96\% AUC scores).

\item Finally, we introduce a technique based on the usage of \textit{synthetic minority oversampling}~\cite{chawla2002smote} to enhance existing datasets by generating additional labeled examples. This will allow us to achieve near perfect classification performance on the account-level bot detection task, by leveraging only a minimal number of features and very small training datasets. 
\end{enumerate}

\subsection{Impact of this work}
A successful tweet-level bot detection approach would potentially overcome the  limitations presented above, namely the need for computationally expensive modes that require large numbers of features, large labeled datasets for training purposes, and access to the recent history of activity of the user profile to scrutinize. 

Given the same pool of users, a tweet-based bot detection approach would have significantly more labeled examples to exploit. For example, in the dataset we use in this paper (discussed in the next section), we have labels for 3,474 human users, which overall generated 8,377,522 tweets; we also have labels for 4,912 social bots, which generated 3,457,344 tweets. 

A tweet-level detection approach would be capable of leveraging nearly 12 million labeled datapoints, while an account-level detection system would only be able to exploit about eight thousands examples of bots and human accounts, while using those millions of tweets to learn patterns associated with the originating accounts. 

Shifting to tweet-level bot detection, and thus having training data orders of magnitude larger than otherwise, makes the problem of bot detection far more amenable to the usage of deep learning models. Such techniques benefit greatly from vast amounts of labeled examples and show extremely high performance in many contexts where such large annotated datasets are available~\cite{lecun2015deep}, from image classification~\cite{krizhevsky2012imagenet} to mastering games~\cite{mnih2013playing,mnih2015human,silver2016mastering}.

Traditional deep learning techniques used for text classification purposes (as well as in the broader context of language models) rely solely on textual features (e.g., characters or n-grams) \citet{john2017survey}. A straightforward implementation of such techniques to tweet-level bot detection could be based exclusively on tweet texts as inputs for the deep neural network of choice.
However, prior results in bot detection suggested that tweet text alone is not highly predictive of bot accounts ~\cite{ferrara2016rise}. Exploiting additional  features such as account metadata, network structure information, or temporal activity patterns, have been found to yield more robust and accurate results. 

To draw a parallel with recent advances in natural language processing (NLP) powered by deep learning, we here propose a novel \textit{Contextual LSTM} architecture that utilizes both tweet text and account metadata (which are provided by the Twitter API alongside with the tweet itself, and do not require extra data collection steps) to yield a high classification accuracy. 

We hypothesize that the proposed model can be used in other deep learning applications where multimodal data are available for such types of classification tasks.


A successful tweet-level bot detection system also has interesting practical implications. 
\begin{itemize}
\item Identifying instances of large numbers of bot-generated tweets coming from a single account would enable us to identify bots that have identifiably bot-generated tweets interspersed with human generated tweets, whether manually generated or retweeted. 
\item Since tweets are often viewed as a part of a feed containing both genuine and bot-generated tweets, it is potentially useful to be able to flag isolated tweets as possibly bot-generated.
\end{itemize}

\section{Dataset}
 
The dataset used in our work is the dataset presented in \cite{cresci2017paradigm}, which contains an entirely new breed of social bots. We use a mixture of the groups \verb genuine  \verb accounts , \verb social  \verb spambots  \verb #1 ,  \verb social  \verb spambots  \verb #2  and   \verb social  \verb spambots  \verb #3 . 

All these subsets of data together have over 8,386 user accounts, and over 11,834,866 tweets to train on. A group-wise breakdown may be seen in Table 1. \par 

\begin{table}[h] 
  \begin{tabular}{ |c|c|c| }
    \hline 
Dataset & Accounts & Tweets \\
  \hline
  \verb genuine  \verb accounts  & 3,474 & 8,377,522   \\ 
  \verb social  \verb spambots  \verb #1  & 991 & 1,610,176  \\ 
  \verb social  \verb spambots  \verb #2  & 3,457 &  428,542  \\ 
  \verb social  \verb spambots  \verb #3  & 464 &  1,418,626 \\
  \hline
  \end{tabular}
  \caption{Breakdown of the dataset used to train our models. The dataset was obtained from Cresci and collaborators~\cite{cresci2017paradigm}.}
  \label{tab:1}\vspace*{-.6cm}
\end{table}

Though many established techniques use a large number of features (\cite{davis2016botornot}, for example uses over 1,500 features), recent research \cite{ferrara2017disinformation,ferrara2016rise} shows that similar high performance can be obtained by using a minimal number of features. For account-level bot detection, we use the following features:
\begin{itemize}
\item Statuses Count
\item Followers Count
\item Friends Count
\item Favorites Count
\item Listed Count
\item Default Profile
\item Geo Enables
\item Profile Uses Background image
\item Verified
\item Protected
\end{itemize}

 Similarly, for tweet level classification we use only 6 features, apart from the tweet content itself:
\begin{itemize}
\item Retweet Count 
\item Reply Count
\item Favorite Count
\item Number of Hashtags
\item Number of URLs
\item Number of Mentions
\end{itemize}

The choice of limiting the size of the feature set is motivated by two important reasons: 

\begin{itemize}
\item \textbf{Model efficiency:} A reduced set of features yields very efficient models that can be trained faster and are less prone to overfitting, which is a common issue in social media data mining due to the presence of outliers.
\item \textbf{Interpretability:} A limited set of features with an obvious meaning, like the ones provided by account metadata, allows to produce interpretable models. This is a very important point, especially when combined with deep learning strategies that are notoriously hard to interpret.
\end{itemize}

Our choice goes in antithesis to that of many feature-based systems, which are designed to leverage hundreds, or even thousands of features, but whose computational efficiency is suboptimal and whose interpretability is challenging. 

\section{Methods}

In this study we face two classification tasks: account-level bot detection and tweet-level bot detection. In the following, we describe the methodological approaches that we adopted to address these two challenges.

\subsection{Task 1: Account-level Classification}

Previous work on account-level classification has found that user metadata tends to be the best predictor for bot detection \cite{ferrara2017disinformation}. As presented in Section 2, we use a minimal number of highly interpretable features that require little to no preprocessing. This enabled us to use a multitude of out-of-the-box classical machine learning approaches as listed in Section 4. 

We found that most of these approaches sufficed, with most approaches crossing AUC $90\%$. Our most successful approach was using an Random Forest classifier, giving an AUC of $98.45\%$. \par 

However, significant performance gains were observed on balancing the dataset with \textit{oversampling techniques}, specifically the synthetic minority oversampling technique (SMOTE)~\cite{chawla2002smote}. 

The SMOTE algorithm generates samples based on the feature space of the minority examples (i.e., the class that has the fewer number of labeled datapoints), and  is a powerful method that has seen successfully across many domains~\cite{he2009learning}. Specifically, we use a combination of SMOTE and two undersampling techniques. Such data enhancement techniques are used to remove any bias introduced by oversampling. Here, we combine SMOTE with data enhancement via (1) \textit{Edited Nearest Neighbors} (ENN) ~\cite{wilson1972asymptotic} and (2) \textit{Tomek Links} ~\cite{tomek1976two}. These two combinations have been found to give excellent results on imbalanced data~\cite{batista2004study}. 

Though combining SMOTE and undersampling through Tomek Links (SMOTOMEK) does not improve results by much, as we will discuss later, significant improvement is seen by combining SMOTE and undersampling through ENN (SMOTENN) across all models. With SMOTENN, near perfect classification accuracy is achieved with the best model being an AdaBoost Classifier at $99.81\%$ accuracy. \par 

Our results will suggest that near-perfect accuracy bot detection prediction at the account level can be achieved even without complex deep learning architectures. The same does not hold for the next task, i.e,  that of detecting bots from single observations.

\subsection{Task 2: Tweet-level Classification}
We here introduce the problem of determining from a single datapoint (e.g., a single tweet) whether the user in question is a bot or not. 

The approaches that do use tweet content tend to use feature engineering and specific characteristics of the tweets, such as those extracted via Parts-of-Speech tagging, counting the number of hashtags or measuring tweet dissimilarity~\cite{ferrara2016rise}. For bot detection in particular, there is a dearth of convincingly successful approaches based on using single observations. \par 

As a baseline, we attempted to use an approach similar to that of Section 3.1 by exploiting just the features described in Section 2. Without oversampling, however, none of our methods exceeded an AUC of $77\%$. By means of oversampling with SMOTE followed by undersampling through ENN, however, our results somewhat improve, as we will discuss in Section 4.1. \par 

Many of the state-of-the-art techniques from NLP using textual content tend to focus on the average pattern of tweeting style. Mostly relying on traditional data mining and NLP techniques, these methods have been proven ineffective against more advanced social bots~\cite{cresci2017paradigm}. To overcome the limitations of traditional techniques, we use Long Short Term Memory (LSTM) models ~\cite{hochreiter1997long}, a superior variant of Recurrent Neural Networks (RNNs) ~\cite{jozefowicz2015empirical}. RNNs and their variants have been found to been effective for NLP tasks, given their ability to learns relationships in sequential data ~\cite{goldberg2016primer}. \par 



To transform the tweets into a form suitable for LSTMs, as an embedding we use a pre-trained set of Global Vectors for Word Representation (GloVE) meant for Twitter data \cite{pennington2014glove}. GloVE is a global log-bilinear regression model that global matrix factorization and local context window methods to effectively learn the substructure of natural language, by training on word co-occurrence.  Prior to using this embedding, we  preprocess our tweets by tokenizing them using the methods suggested by the creators of GloVE, as follows.

\subsubsection{Preprocessing Data}

Prior to training the LSTM on the tweets, we preprocess the data by forming a string of tokens from each tweet.
	\begin{itemize}
	\item We replace occurrences of hashtags, URLs, numbers and user mentions with the tags ``$<$hashtag$>$", ``$<$url$>$", ``$<$num- ber$>$", or `$<$user$>$".
    \item Similarly, most common emojis are replaced with ``$<$smile$>$", ``$<$heart$>$",  ``$<$lolface$>$", ``$<$neutralface$>$" or ``$<$angryface$>$", depending on the specific emoji.  
    \item For words written in upper case letters or for words containing more than 2 repeated letters, a tag denoting that is placed after the occurrence of the word. For example, the word ``HAPPY" would by replaced by two tokens, ``happy" and ``$<$allcaps$>$". 
    \item All tokens are converted to lower case. 
	\end{itemize}

Then, these tokenized tweets are  transformed into an embedding using the aforementioned pre-trained GloVE model. The resulting sequence of vectors is then fed to the LSTM that outputs a single 32-dimension vector that is then fed forward through 2 ReLU activated layers of size 128 and 64 to give the output. A diagram of this simple LSTM architecture can be seen in Figure 1. It is to be noted that our model resets its state after each input, and therefore only learning sequential structure within each tweet and not the sequence of tweets.  \par

\begin{figure}
\begin{center}
\includegraphics[width=0.4\textwidth]{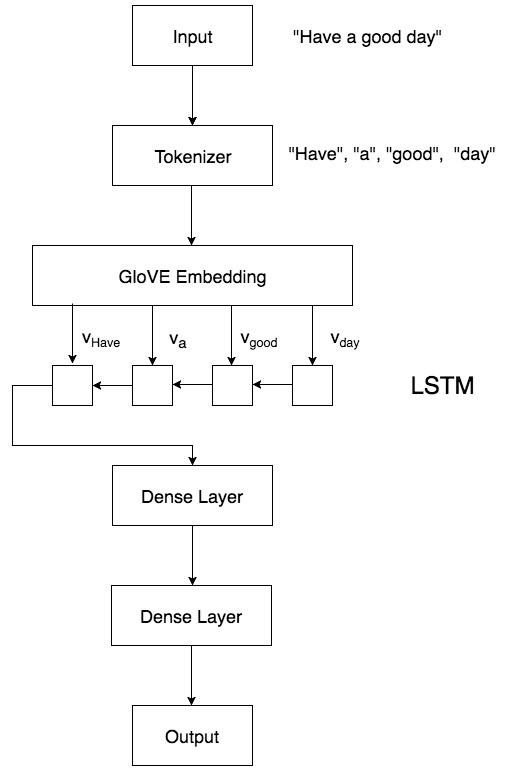} 
\caption{Architecture of model for tweet-level bot detection that takes only the tweet content as its input.}
\end{center}\vspace*{-.6cm}
\end{figure}

\subsubsection{Limits of traditional LSTM deep nets}

However, this approach only uses the text content of the tweets, and does not utilize the metadata associated with it. As observed from the results of Table 3, which will be discussed later into detail, although using metadata does not precisely predict the nature of a user, it does at least weakly predict the same. In other words, metadata are weak predictors of an account's nature (bot or not). 

Since metadata are not sequential data (unlike tweet text) that can be fed into the LSTM along with the text embedding, we instead propose a new \textit{Contextual LSTM} architecture that can effectively utilize both the text and the metadata, even though they are predictors of different strengths. \par

\subsubsection{Contextual LSTM architecture}
As seen in Figure 2, our proposed architecture has multiple inputs and outputs. Multi-input recurrent models have been suggested before, mainly in language models: for example, some authors~\cite{hoang2016incorporating,auli2013joint,mikolov2012context} adopt the idea of giving auxiliary inputs to either the input, hidden, or output layer of the recurrent model to enrich the learned representations. 

In our work, we give auxiliary input to the output layer. Similar to the tweets-only model previously described, the main input is the tweet text that is tokenized and transformed into a set of GloVE vectors before feeding them into the LSTM. This again results in an output vector that is concatenated with our auxiliary input and then given as input to a 2-layer neural network with ReLU activations to yield the output. The exact sizes of layers are the same as that of our previous model. 

As a regularization mechanism, we introduce an auxiliary output after the LSTM output whose target is also the classification label. Such a mechanism has been used successfully before in \cite{szegedy2015going}. The total loss is a weighted average of the auxiliary output loss and the main output loss (0.2:0.8 in this case).

\begin{figure}
\begin{center}
\includegraphics[width=0.4\textwidth]{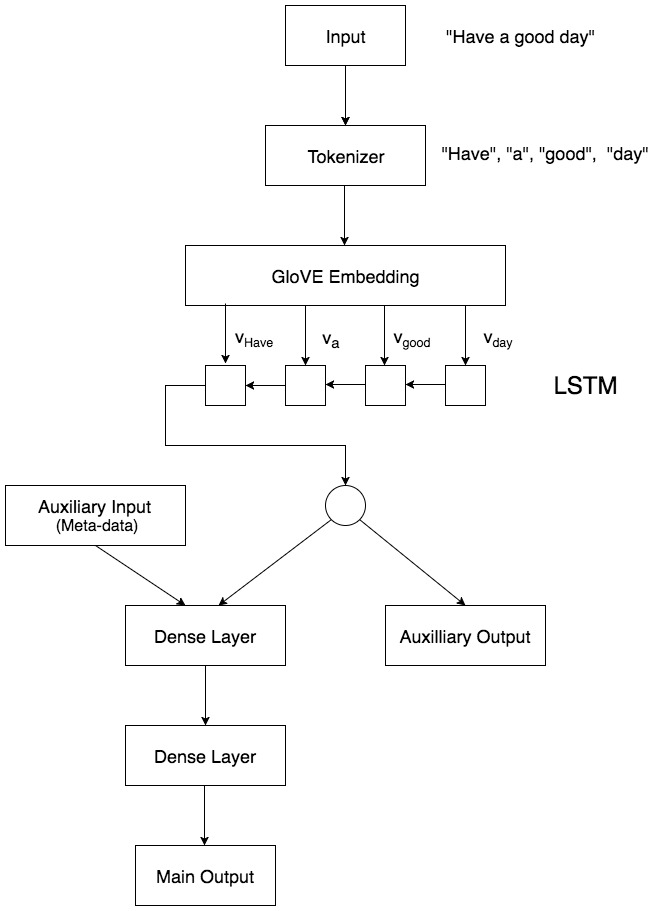} 
\caption{Architecture of the proposed contextual LSTM that uses both tweet content and the metadata that comes with the tweet.}
\end{center}\vspace*{-.6cm}
\end{figure}

\section{Results}

Here, we have tabulated our results for the approaches outlined in the previous section. The success of each method is measured on a variety of metrics, namely precision, recall, f1-score, accuracy and Area Under the Receiver Operating Characteristic Curve (AUC/ROC).

\subsection{Task 1: Account-Level Classification}

In Table 2, we have tabulated results of our experiments on account-level classification. 

The first batch of five classifiers shows the results by solely using the data without any oversampling methods: Random Forest Classifiers and AdaBoost Classifiers yield the best results, with an AUC $> 98\%$. Overall, all classifiers perform really well, suggesting that the account-level classification task, at least for the dataset under analysis here, is not particularly daunting. \par 

The second batch of systems reported in Table 2 shows the result obtained after using  oversampling with SMOTE, followed by undersampling with ENN (SMOTENN): the results are improved across all classifiers, with the AdaBoost classifier achieving near perfect accuracy of $99.81\%$. This suggests that our strategy of synthetic minority oversampling is really effective when dealing with this types of account-level classification tasks. 

The third batch of systems shows the results of oversampling with SMOTE followed by undersampling with TOMEK (SMOTOMEK): with this approach  the results improve as well, though not as drastically as by using ENN. \par 

Overall, we demonstrated the feasibility of high-accuracy account-level bot detection by means of unsophisticated off-the-shelf algorithms in combination with synthetic minority oversampling techniques to enhance training data.

\begin{table*}[t]\small
  \centering
  \begin{tabular}{ |c|c|c|c|c|c| }
    \hline 
System & Precision & Recall & F1-Score & Accuracy & AUC/ROC \\
  \hline
  Logistic Regression & 0.94 & 0.93 & 0.93 & 0.9066 & 0.8891 \\ 
  SGD Classifier & 0.87 & 0.87 & 0.87 & 0.8726 & 0.8680 \\ 
  Random Forest Classifier & 0.98 & 0.98 & 0.98 & \textbf{0.9839} & \textbf{0.9845} \\ 
  AdaBoost Classifier & 0.98 & 0.98 & 0.98 & 0.9823 & 0.9823 \\ 
  2-layer NN (500,200,1) RelU$+$Adam & 0.95 & 0.95 & 0.95 & 0.9496 & 0.9475 \\ \hline 
  Logistic Regression  (With SMOTENN) & 0.99 & 0.99 & 0.99 & 0.9859 & 0.9862 \\ 
SGD Classifier (With SMOTENN) & 0.95 & 0.94 & 0.94 & 0.9433 & 0.9443 \\ 
Random Forest Classifier (With SMOTENN) & 0.99 & 0.99 & 0.99 & 0.9937 & 0.9938 \\ 
AdaBoost Classifier (With SMOTENN)  & 1.00 & 1.00 & 1.00 & \textbf{0.9981} & \textbf{0.9981} \\ 
2-layer NN (300,200,1) RelU$+$Adam (With SMOTENN) & 0.99 & 0.99 & 0.99 & 0.9878 & 0.9879 \\ \hline 
  Logistic Regression (With SMOTOMEK) & 0.92 & 0.91 & 0.91 & 0.9094 & 0.9098 \\ 
SGD Classifier (With SMOTOMEK) & 0.90 & 0.90 & 0.90 & 0.9030 & 0.9031 \\ 
Random Forest Classifier (With SMOTOMEK) & 0.99 & 0.99 & 0.99 & 0.9859 & 0.9859 \\ 
AdaBoost Classifier (With SMOTOMEK) & 0.99 & 0.99 & 0.99 & \textbf{0.9865} & \textbf{0.9865} \\ 
2-layer NN (300,200,1) RelU$+$Adam (With SMOTOMEK) & 0.95 & 0.95 & 0.95 & 0.9391 & 0.9489 \\ 
\hline
  \end{tabular}
  \caption{Classification performance of various systems on the account-level (user) bot detection task. The first batch of systems represent traditional off-the-shelf baseline approaches, that already exhibit very accurate performance. The second and third batches of systems are enhanced by means of synthetic minority oversampling techniques, to illustrate how it is possible to achieve nearly perfect account-level bot detection without the need for complex deep architectures. For each batch of systems we highlighted the best accuracy and AUC/ROC performing ones: AdaBoost consistently provides the top (or nearly the top) performance across all account-level bot detection benchmarks. }
  \label{tab:1}\vspace*{-.6cm}
\end{table*}

\subsection{Task 2: Tweet-Level Classification}

We now illustrate the proposed tweet-level bot detection task. 
Table 3 shows four batches of systems: the first three show the baseline models provided by the off-the-shelf classifiers, specifically in their naive implementation (first batch), and by using again synthetic minority oversampling in combination with ENN (second batch) or TOMEK (third batch); finally, the fourth and last batch shows the performance of the proposed LSTM deep architectures, showing different configurations.

The first batch of systems that only uses the tweet metadata does not work particularly well. Without data augmentation, none of the results cross an AUC of 78\%. 

By using synthetic minority oversampling, in combination with ENN, results drastically improve: the second batch of systems shows that all models approach an accuracy between 88\% and 90\%. 

However, no such improvement is seen by using SMOTE in combination with TOMEK. Although the reasons of such a difference with respect to the previous task are not apparent, and warrant further investigation, we hypothesized that ENN works better because tweets originated by bots exhibit identical or very similar metadata thus a nearest-neighbor strategy works well. \par 

The fourth batch of results present our architecture and deserves an in-depth discussion.
By using only the tweets, our LSTM system provides a classification accuracy of $95.53\%$ (\textit{cf.} ``LSTM (Tweets-only + 50D GloVE)''. This result is very promising: the use of our architecture yields a lift in performance in the order of 5\% even just using the tweet texts. 

Following the intuition that metadata can enrich the information available for classification purposes, the last four systems show the results of our \textit{Contextual LSTM} combining metadata and tweet text features. Our Contextual LSTM shows a boost in performance yielding a classification accuracy of about $96.33\%$ for the best model with 200-dimension GloVE embedding (\textit{cf.} ``Contextual LSTM (200D GloVE)''). It is worth noting different configuration of GloVE, namely varying the dimensional of the word embedding space, do not affect the performance much: a general trend seems to suggest that higher dimensionality yield increasingly slightly better performance.
In conclusion, from our analysis we derived that, even though metadata is shown to be a weak predictor for our baseline results, our proposed architecture uses the extra information provided by the metadata to yield slightly more accurate prediction results. It is to be noted that the metadata has not been oversampled for the Contextual LSTM systems. \par

\begin{table*}[t]\small
  \centering
  \begin{tabular}{ |c|c|c|c|c|c| }
    \hline 
System & Precision & Recall & F1-Score & Accuracy & AUC/ROC \\
  \hline
  Logistic Regression (Metadata-only)& 0.80 & 0.80 & 0.79 & 0.8008 & 0.7633 \\ 
SGD Classifier (Metadata-only) & 0.76 & 0.76 & 0.75 & 0.7625 & 0.7191 \\ 
Random Forest Classifier (Metadata-only)& 0.80 & 0.80 & 0.80 & \textbf{0.8042} & \textbf{0.7765} \\ 
AdaBoost Classifier (Metadata-only)& 0.80 & 0.80 & 0.79 & 0.7991 & 0.7618 \\ \hline
Logistic Regression (Metadata-only + SMOTENN) & 0.92 & 0.92 & 0.92 & 0.9188 & 0.8820 \\ 
SGD Classifier (Metadata-only + SMOTENN) & 0.91 & 0.90 & 0.90 & 0.8992 & 0.8860 \\ 
Random Forest Classifier (Metadata-only + SMOTENN) & 0.92 & 0.92 & 0.92 & 0.9233 & 0.8806 \\ 
AdaBoost Classifier (Metadata-only + SMOTENN) & 0.93 & 0.92 & 0.93 & \textbf{0.9234} & \textbf{0.9065} \\ \hline 
Logistic Regression (Metadata-only+SMOTOMEK)& 0.79 & 0.77 & 0.76 & 0.7666 & 0.7667 \\ 
SGD Classifier (Metadata-only+SMOTOMEK)& 0.78 & 0.77 & 0.76 & 0.7664 & 0.7664 \\ 
Random Forest Classifier (Metadata-only+SMOTOMEK)& 0.79 & 0.77 & 0.77 & \textbf{0.7747} & \textbf{0.7748} \\ 
AdaBoost Classifier (Metadata-only+SMOTOMEK)& 0.79 & 0.77 & 0.77 & 0.7715 & 0.7716 \\ \hline 
LSTM (Tweet-only + 50D GloVE) & 0.96 & 0.96 & 0.96 & 0.9553 & 0.9567 \\ 
Contextual LSTM (25D GloVE) & 0.96 & 0.96 & 0.96 & 0.9567 & 0.9585 \\ 
Contextual LSTM (50D GloVE) & 0.96 & 0.96 & 0.96 & 0.9618 & 0.9627 \\ 
Contextual LSTM (100D GloVE) & 0.96 & 0.96 & 0.96 & 0.9618 & 0.9626 \\ 
Contextual LSTM (200D GloVE) & 0.96 & 0.96 & 0.96 & \textbf{0.9633} & \textbf{0.9643} \\ 
\hline
  \end{tabular}
  \caption{Classification performance of various systems, including the proposed Contextual LSTM, on the tweet-level bot detection task. The first batch of systems represent traditional off-the-shelf baseline approaches: their accuracy and AUC/ROC scores range between 71\% and 80\%. The second and third batches of systems are enhanced by means of synthetic minority oversampling techniques, showing better classification performance between 76\% and 90\% accuracy and AUC/ROC scores. The LSTM architecture that we propose is presented in the forth batch of systems: our results outperform the other approaches by a significant margin, averaging above 96\% accuracy and AUC/ROC scores, to demonstrate that tweet-level bot detection can be achieved with extremely high accuracy, small number of features, and limited-size training datasets. For each batch of systems we highlighted the best Accuracy and AUC/ROC performing ones. Among the traditional machine learning models (i.e., excluding the proposed approach that significantly outperforms all the baselines and their variants using synthetic minority oversampling), both Random Forest and AdaBoost classifiers appear to consistently deliver the best performance across all tweet-level bot detection benchmarks.}
  \label{tab:1}\vspace*{-.6cm}
\end{table*}

\begin{figure*}
\begin{center}
\includegraphics[width=\textwidth,clip=true,trim=0 15 0 10]{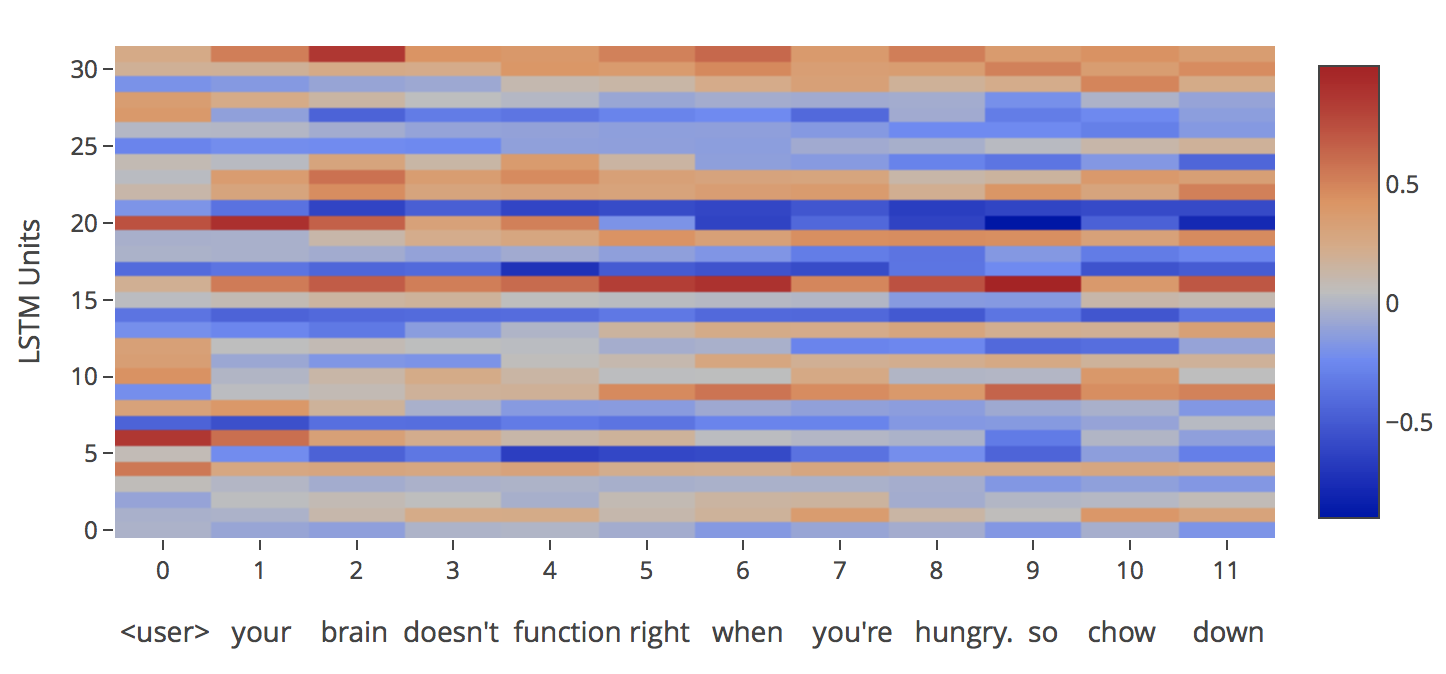} 
\caption{Representations over time of LSTM 32 hidden units activations for a tweet generated by a human. Each column corresponds to LSTM outputs at each time step. Cells correspond to the 32 dimensions of the representation at each timestep.}\label{fig:geneg}
\end{center}
\end{figure*}

\begin{figure*}
\begin{center}
\includegraphics[width=\textwidth,clip=true,trim=0 15 0 10]{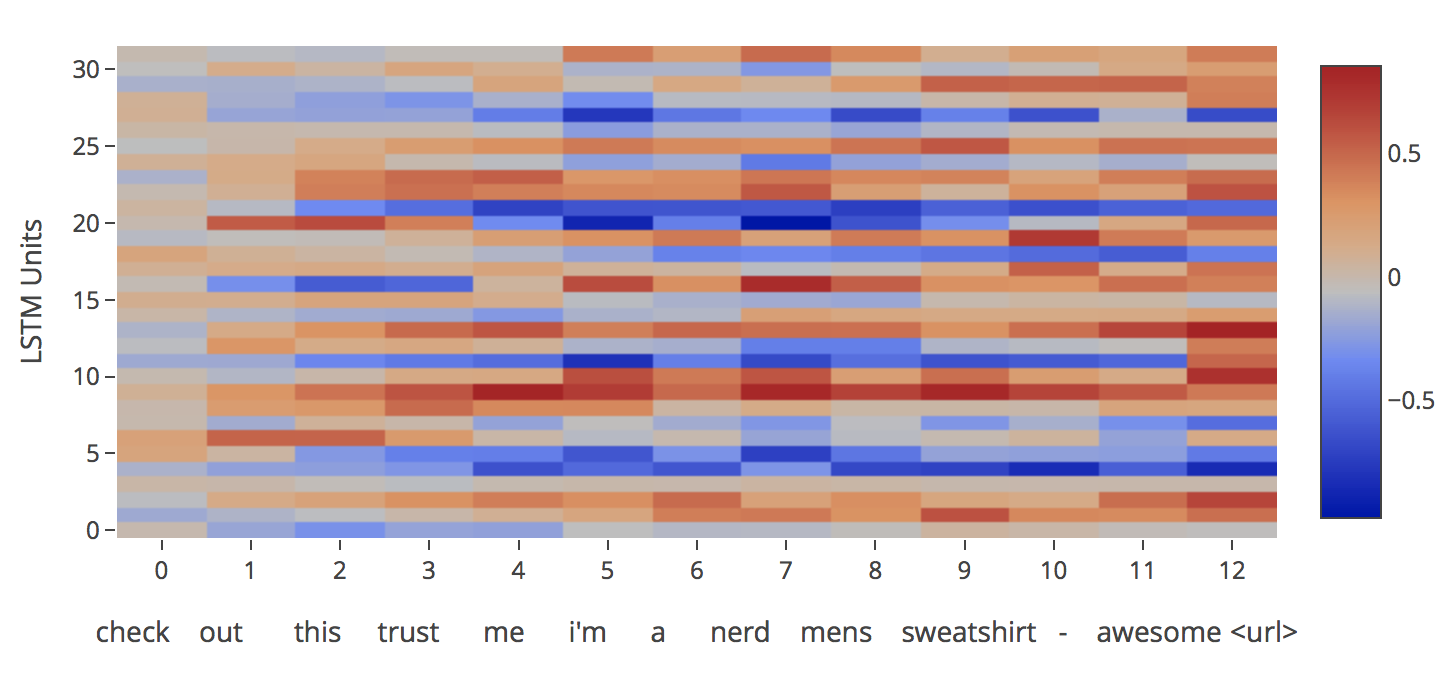} 
\caption{Representations over time of LSTM 32 hidden units activations for a tweet  generated by a bot. Each column corresponds to LSTM outputs at each time step. Cells correspond to the 32 dimensions of the representation at each timestep.}\label{fig:boteg}
\end{center}
\end{figure*}

\begin{figure*}[t]
\begin{center}
\includegraphics[width=\textwidth,clip=true,trim=0 10 0 10]{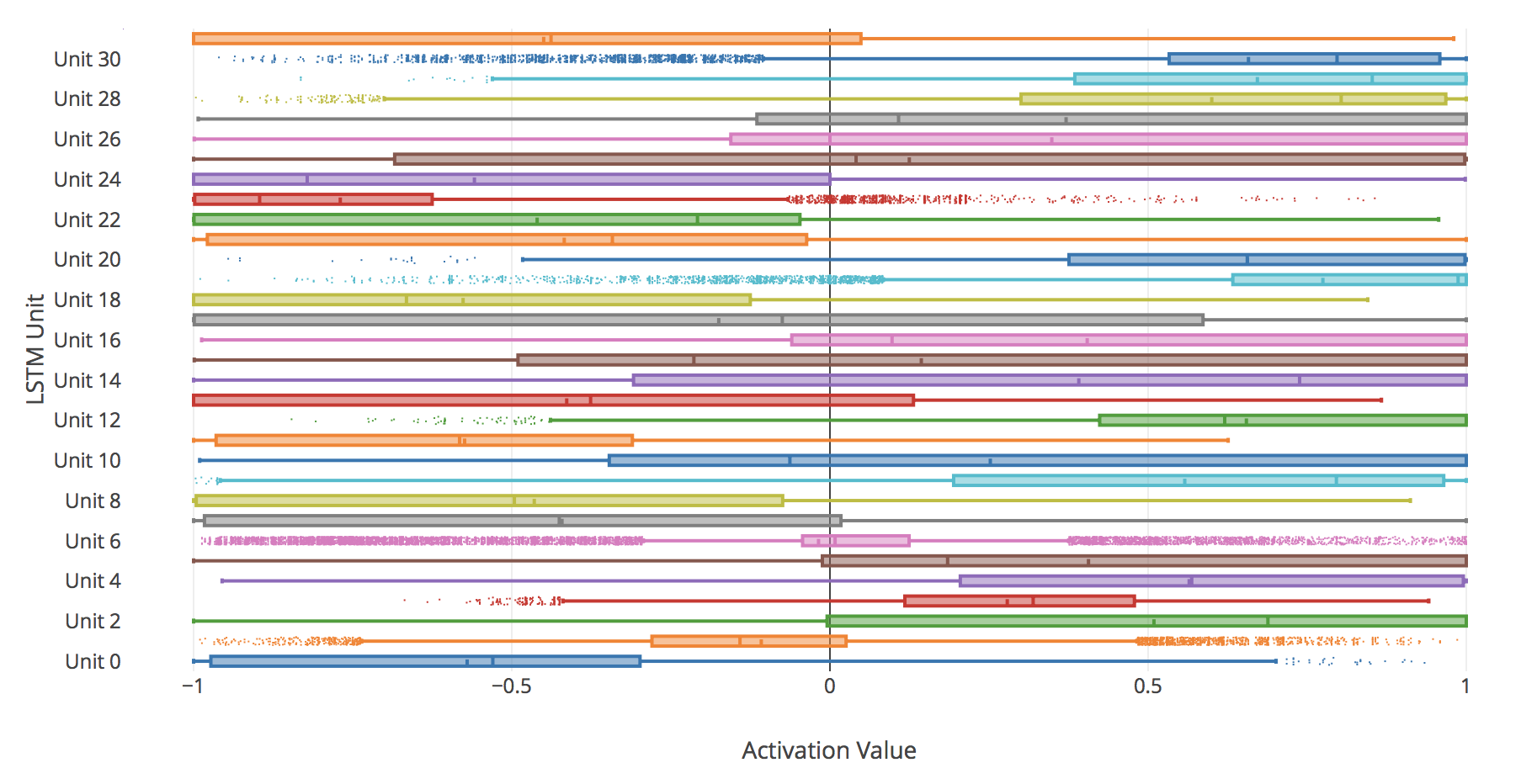} 
\caption{Distribution of activation values of LSTM hidden units (of which there are 32) on tweets generated by bots.}\label{fig:gendist}
\end{center}
\end{figure*}

\begin{figure*}[t]
\begin{center}
\includegraphics[width=\textwidth,clip=true,trim=0 10 0 10]{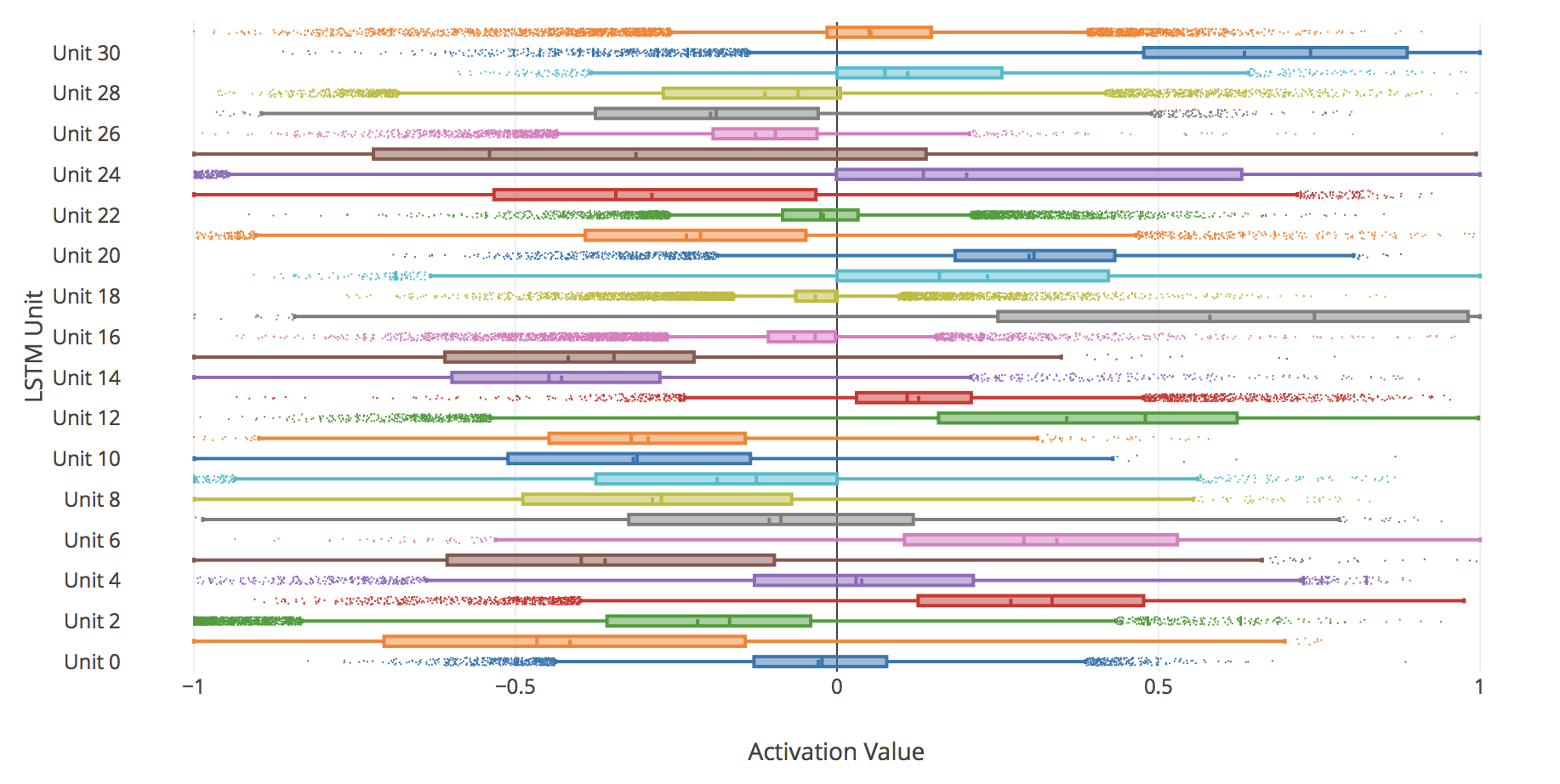} 
\caption{Distribution of activation values of LSTM hidden units (of which there are 32) on tweets generated by genuine users.}\label{fig:botdist}
\end{center}
\end{figure*}

\subsection{Analysis}

Deep neural network models are often touted as uninterpretable black boxes. Although they provide impressive results in many practical applications, various researchers, practitioners, policy makers, and funding agencies~\cite{socialsim} demand attention to the matter of interpretability~\cite{samek2017explainable}. Recently,  several studies aimed at shedding light on the inner states of recurrent models by visualizing their hidden state dynamics. Karpathy and collaborators, for example, examined inner states of language models~\cite{karpathy2015visualizing}, while Li \textit{et al.}~\cite{li2015visualizing} introduced strategies to help interpreting deep neural network results in NLP tasks. With this in mind, we here attempt to understand and interpret the effectiveness of our methods by analyzing the hidden state activations of our neural network model. \par  

We first visualize the change in the representation, i.e., the output over time of the LSTM hidden units. We give examples of tweets generated by a human and by a bot in Figures \ref{fig:geneg} and \ref{fig:boteg} respectively. We observe significant differences in the activations between the examples, suggesting that the LSTM hidden units may correspond to different linguistic features. \par


It has been observed by Cresci \textit{et al.} ~\cite{cresci2017paradigm} that using simple textual features, as seen in prior work~\cite{miller2014twitter},  reveals ineffective against the more sophisticated social bots we study in this work. We plot the distributions of the final LSTM representations of genuine tweets and bot-generated tweets in Figures \ref{fig:gendist} and \ref{fig:botdist} respectively. We see that the majority of hidden units have a significant difference in the distribution of activation values for genuine tweets versus to bot-generated tweets. For example, Unit 0 (bottom blue box) has a broad activation range centered around -0.6 for tweets generated by bots, as opposed to a narrow activation range centered around 0 for tweets generated by humans. Similarly, Unit 31 (top orange box) is activated in the negative [-1,0] range for bot-generated tweets, but again narrowly around 0 for genuine tweets. These examples suggest the mechanism behind the ability of LSTM to learn more complex, non-linear features, which are effective in identifying bot-generated tweets even from single observations. \par

\subsection{Considerations and Limitations}

We developed bot detection systems requiring a minimal number of features as well as one single observation of a tweet generated by the account to be scrutinized. For the account-level bot detection, we showed that using oversampling and data enhancement techniques could yield models that perform nearly perfectly. 

Our main contribution, however, is presenting to the best of our knowledge the first bot detection model that detects the nature of an account from a single tweet. First, we find that simply using the tweet metadata does not suffice. We also show that while simply using the content of the tweet works quite well with our proposed LSTM architecture, we can slightly improve the performance utilizing the information from the metadata. \par 


\section{Related Work}
Bots (a.k.a., social bots, or \textit{sybil accounts}) have been found guilty of polluting social media conversations in a variety of scenarios. Reports of online manipulation mediated by bots span political conversation~\cite{metaxas2012social,forelle2015political,howard2016bots,woolley2016automating,bessi2016social}, fake news~\cite{ferrara2017disinformation,shao2017spread}, conspiracy theories~\cite{subrahmanian2016darpa}, stock market manipulation~\citep{ferrara2015manipulation}, public health~\cite{clark2015vaporous}, and more~\cite{haustein2016tweets} --- it is worth noting that in some rare occasions bots have been used to deliver positive interventions \cite{savage2016botivist,monsted2017evidence}, rather than for nefarious purposes.
The research community promptly responded to the problem of the increasing pervasiveness of bots in platforms like Twitter and Facebook. A wealth of strategies and frameworks have been proposed to address the challenge of bot detection in these environments. A recent review~\cite{ferrara2016rise} proposed a taxonomy describing three types of approaches: \emph{(a)} methods based on social network; \emph{(b)} systems based on crowd-sourcing and human computation; \emph{(c)} algorithms based on predictive features that separate bots from humans. Our framework falls in the last category. 

Bots exhibit great variability and diversity in terms of behavior, capabilities, and intent: this was illustrated with a categorization scheme proposed in another recent survey~\cite{mitter2014categorization}. Another recent white paper discussed the capabilities of bots powered by sophisticated Artificial Intelligence~\cite{adams2017ai}. Bots also attracted the attention of the cyber security research community: Sometimes, large groups of bots are controlled by the same entity, called \textit{bot master}, acting behind the scenes in a command-and-control fashion, in analogy to traditional \textit{botnets} used to deploy cyber attacks and other cyber-security threats~\cite{gu2008botsniffer}, as demonstrated on Twitter as well~\cite{abokhodair2015dissecting,zhou2017starwars}. 

Much work on bot detection assumes extensive access to social media data. For example, Wang and collaborators used clustering techniques to identify large-scale behavioral anomalies~\cite{wang2013you}, while other authors adopted supervised learning to analyze all accounts of some platform and separate bots from humans ~\cite{lee2011seven,beutel2013copycatch,elyashar2013homing,yang2014uncovering}. Although these approaches can be useful, for example to detect large-scale bot infiltrations, they can be implemented exclusively by social media service providers with full access to data and systems. Some of them  published studies showing the effectiveness of some implementations, e.g., SybilRank~\cite{cao2012aiding}, the Facebook Immune System~\cite{stein2011facebook}, and others~\cite{alvisi2013sok}.  
To obviate the limitation of unlimited data access, other techniques have been designed to require smaller samples of user activity, and fewer labeled examples of bot and human users. Examples of such trend include the classification system proposed by Chu \textit{et al.}~\cite{chu2010tweeting,chu2012detecting}, the system based on crowd-sourcing designed by Wang \textit{et al.}~\cite{wang2012social}, the detection techniques based on NLP presented by 
Clark \textit{et al.}~\cite{clark2016sifting}, and BotOrNot~\cite{davis2016botornot}. 

To allow for bot detection at the user level, all these methods still require the analysis of some historical user data, either by indirect data collection ~\cite{chu2010tweeting,chu2012detecting,wang2012social,clark2016sifting}, or, like in the case of BotOrNot~\cite{davis2016botornot}, by interrogating the Twitter API (which imposes strict rate limits, making it impossible to do large-scale bot detection). To the best of our knowledge, no tweet-based detection system existed prior to this work. We filled this gap by designing a LSTM deep neural network based on combinations of textual features and user metadata that is capable of determining if a single tweet is being posted by a human or a bot with extremely high accuracy. The same architecture shows nearly-perfect accuracy in the user-level bot detection task.

\section{Conclusions}


Given the prevalence of sophisticated bots on social media platforms such as Twitter, the need for improved, inexpensive bot detection methods is apparent. We proposed a novel contextual LSTM architecture allowing us to use both tweet content and metadata to detect bots at the tweet level. From a single tweet, our model can achieve an extremely high accuracy exceeding $96\%$ AUC. 

We show that the additional metadata information, though a weak predictor of the nature of a Twitter account per se, when exploited by LSTM decreases the error rate by nearly $20\%$. In addition to this, we propose methods based on synthetic minority oversampling that yield a near perfect user-level detection accuracy ($>99\%$ AUC). Both these methods use a very minimal number of features that can be obtained in a straightforward way from the tweet itself and its metadata, while  surpassing prior state of the art. 

In the future, we plan to make our system open source, and to implement a Web service (for example, an API) to allow the research community to perform tweet-level bot detection using it. From a research standpoint, we plan to use the proposed framework to scrutinize social media conversation in different contexts, in order to determine the extent of the interference of bots with public discourse, as well as to understand how their capabilities and sophistication evolve over time.

\begin{acks}
\footnotesize{The authors gratefully acknowledge support by the Air Force Office of Scientific Research (AFOSR \#FA9550-17-1-0327), and by the Defense Advanced Research Projects Agency (DARPA \#W911NF-17-C-0094). The U.S. Government is authorized to reproduce and distribute reprints for Governmental purposes notwithstanding any copyright annotation thereon. The views and conclusions contained herein are those of the authors and should not be interpreted as necessarily representing the official policies or endorsements, either expressed or implied, of AFOSR, DARPA, or the U.S. Government.}
\end{acks}


\balance
\bibliographystyle{ACM-Reference-Format}
\bibliography{bibliography}


\begin{thebibliography}{00}


\ifx \showCODEN    \undefined \def \showCODEN     #1{\unskip}     \fi
\ifx \showDOI      \undefined \def \showDOI       #1{#1}\fi
\ifx \showISBNx    \undefined \def \showISBNx     #1{\unskip}     \fi
\ifx \showISBNxiii \undefined \def \showISBNxiii  #1{\unskip}     \fi
\ifx \showISSN     \undefined \def \showISSN      #1{\unskip}     \fi
\ifx \showLCCN     \undefined \def \showLCCN      #1{\unskip}     \fi
\ifx \shownote     \undefined \def \shownote      #1{#1}          \fi
\ifx \showarticletitle \undefined \def \showarticletitle #1{#1}   \fi
\ifx \showURL      \undefined \def \showURL       {\relax}        \fi
\providecommand\bibfield[2]{#2}
\providecommand\bibinfo[2]{#2}
\providecommand\natexlab[1]{#1}
\providecommand\showeprint[2][]{arXiv:#2}

\bibitem[\protect\citeauthoryear{Abokhodair, Yoo, and McDonald}{Abokhodair
  et~al\mbox{.}}{2015}]%
        {abokhodair2015dissecting}
\bibfield{author}{\bibinfo{person}{Norah Abokhodair}, \bibinfo{person}{Daisy
  Yoo}, {and} \bibinfo{person}{David~W McDonald}.}
  \bibinfo{year}{2015}\natexlab{}.
\newblock \showarticletitle{Dissecting a social botnet: Growth, content and
  influence in Twitter}. In \bibinfo{booktitle}{{\em Proc. of the 18th ACM
  Conf. on Computer Supported Cooperative Work \& Social Computing}}. ACM,
  \bibinfo{pages}{839--851}.
\newblock


\bibitem[\protect\citeauthoryear{Adams}{Adams}{2017}]%
        {adams2017ai}
\bibfield{author}{\bibinfo{person}{Terrence Adams}.}
  \bibinfo{year}{2017}\natexlab{}.
\newblock \showarticletitle{AI-Powered Social Bots}.
\newblock \bibinfo{journal}{{\em arXiv preprint arXiv:1706.05143\/}}
  (\bibinfo{year}{2017}).
\newblock


\bibitem[\protect\citeauthoryear{Alvisi, Clement, Epasto, Lattanzi, and
  Panconesi}{Alvisi et~al\mbox{.}}{2013}]%
        {alvisi2013sok}
\bibfield{author}{\bibinfo{person}{Lorenzo Alvisi}, \bibinfo{person}{Allen
  Clement}, \bibinfo{person}{Alessandro Epasto}, \bibinfo{person}{Silvio
  Lattanzi}, {and} \bibinfo{person}{Alessandro Panconesi}.}
  \bibinfo{year}{2013}\natexlab{}.
\newblock \showarticletitle{Sok: The evolution of sybil defense via social
  networks}. In \bibinfo{booktitle}{{\em Proc. IEEE Symposium on Security and
  Privacy (SP)}}. \bibinfo{pages}{382--396}.
\newblock


\bibitem[\protect\citeauthoryear{Auli, Galley, Quirk, and Zweig}{Auli
  et~al\mbox{.}}{2013}]%
        {auli2013joint}
\bibfield{author}{\bibinfo{person}{Michael Auli}, \bibinfo{person}{Michel
  Galley}, \bibinfo{person}{Chris Quirk}, {and} \bibinfo{person}{Geoffrey
  Zweig}.} \bibinfo{year}{2013}\natexlab{}.
\newblock \showarticletitle{Joint Language and Translation Modeling with
  Recurrent Neural Networks.}. In \bibinfo{booktitle}{{\em EMNLP}},
  Vol.~\bibinfo{volume}{3}. \bibinfo{pages}{0}.
\newblock


\bibitem[\protect\citeauthoryear{Batista, Prati, and Monard}{Batista
  et~al\mbox{.}}{2004}]%
        {batista2004study}
\bibfield{author}{\bibinfo{person}{Gustavo~EAPA Batista},
  \bibinfo{person}{Ronaldo~C Prati}, {and} \bibinfo{person}{Maria~Carolina
  Monard}.} \bibinfo{year}{2004}\natexlab{}.
\newblock \showarticletitle{A study of the behavior of several methods for
  balancing machine learning training data}.
\newblock \bibinfo{journal}{{\em ACM Sigkdd Explorations Newsletter\/}}
  \bibinfo{volume}{6}, \bibinfo{number}{1} (\bibinfo{year}{2004}),
  \bibinfo{pages}{20--29}.
\newblock


\bibitem[\protect\citeauthoryear{Bessi and Ferrara}{Bessi and Ferrara}{2016}]%
        {bessi2016social}
\bibfield{author}{\bibinfo{person}{Alessandro Bessi} {and}
  \bibinfo{person}{Emilio Ferrara}.} \bibinfo{year}{2016}\natexlab{}.
\newblock \showarticletitle{Social bots distort the 2016 US Presidential
  election online discussion}.
\newblock \bibinfo{journal}{{\em First Monday\/}} \bibinfo{volume}{21},
  \bibinfo{number}{11} (\bibinfo{year}{2016}).
\newblock


\bibitem[\protect\citeauthoryear{Beutel, Xu, Guruswami, Palow, and
  Faloutsos}{Beutel et~al\mbox{.}}{2013}]%
        {beutel2013copycatch}
\bibfield{author}{\bibinfo{person}{Alex Beutel}, \bibinfo{person}{Wanhong Xu},
  \bibinfo{person}{Venkatesan Guruswami}, \bibinfo{person}{Christopher Palow},
  {and} \bibinfo{person}{Christos Faloutsos}.} \bibinfo{year}{2013}\natexlab{}.
\newblock \showarticletitle{Copycatch: stopping group attacks by spotting
  lockstep behavior in social networks}. In \bibinfo{booktitle}{{\em Prov. 22nd
  Intl. ACM Conf. World Wide Web (WWW)}}. \bibinfo{pages}{119--130}.
\newblock


\bibitem[\protect\citeauthoryear{Cao, Sirivianos, Yang, and Pregueiro}{Cao
  et~al\mbox{.}}{2012}]%
        {cao2012aiding}
\bibfield{author}{\bibinfo{person}{Qiang Cao}, \bibinfo{person}{Michael
  Sirivianos}, \bibinfo{person}{Xiaowei Yang}, {and} \bibinfo{person}{Tiago
  Pregueiro}.} \bibinfo{year}{2012}\natexlab{}.
\newblock \showarticletitle{Aiding the detection of fake accounts in large
  scale social online services}. In \bibinfo{booktitle}{{\em 9th USENIX Symp on
  Netw Sys Design \& Implement}}. \bibinfo{pages}{197--210}.
\newblock


\bibitem[\protect\citeauthoryear{Chawla, Bowyer, Hall, and Kegelmeyer}{Chawla
  et~al\mbox{.}}{2002}]%
        {chawla2002smote}
\bibfield{author}{\bibinfo{person}{Nitesh~V Chawla}, \bibinfo{person}{Kevin~W
  Bowyer}, \bibinfo{person}{Lawrence~O Hall}, {and} \bibinfo{person}{W~Philip
  Kegelmeyer}.} \bibinfo{year}{2002}\natexlab{}.
\newblock \showarticletitle{SMOTE: synthetic minority over-sampling technique}.
\newblock \bibinfo{journal}{{\em Journal of artificial intelligence
  research\/}}  \bibinfo{volume}{16} (\bibinfo{year}{2002}),
  \bibinfo{pages}{321--357}.
\newblock


\bibitem[\protect\citeauthoryear{Chu, Gianvecchio, Wang, and Jajodia}{Chu
  et~al\mbox{.}}{2010}]%
        {chu2010tweeting}
\bibfield{author}{\bibinfo{person}{Zi Chu}, \bibinfo{person}{Steven
  Gianvecchio}, \bibinfo{person}{Haining Wang}, {and} \bibinfo{person}{Sushil
  Jajodia}.} \bibinfo{year}{2010}\natexlab{}.
\newblock \showarticletitle{Who is tweeting on Twitter: human, bot, or
  cyborg?}. In \bibinfo{booktitle}{{\em Proc. 26th annual computer security
  applications conf.}} \bibinfo{pages}{21--30}.
\newblock


\bibitem[\protect\citeauthoryear{Chu, Gianvecchio, Wang, and Jajodia}{Chu
  et~al\mbox{.}}{2012}]%
        {chu2012detecting}
\bibfield{author}{\bibinfo{person}{Zi Chu}, \bibinfo{person}{Steven
  Gianvecchio}, \bibinfo{person}{Haining Wang}, {and} \bibinfo{person}{Sushil
  Jajodia}.} \bibinfo{year}{2012}\natexlab{}.
\newblock \showarticletitle{Detecting automation of twitter accounts: Are you a
  human, bot, or cyborg?}
\newblock \bibinfo{journal}{{\em IEEE Tran Dependable \& Secure Comput\/}}
  \bibinfo{volume}{9}, \bibinfo{number}{6} (\bibinfo{year}{2012}),
  \bibinfo{pages}{811--824}.
\newblock


\bibitem[\protect\citeauthoryear{Clark, Jones, Williams, Kurti, Nortotsky,
  Danforth, and Dodds}{Clark et~al\mbox{.}}{2015}]%
        {clark2015vaporous}
\bibfield{author}{\bibinfo{person}{Eric Clark}, \bibinfo{person}{Chris Jones},
  \bibinfo{person}{Jake Williams}, \bibinfo{person}{Allison Kurti},
  \bibinfo{person}{Michell Nortotsky}, \bibinfo{person}{Christopher Danforth},
  {and} \bibinfo{person}{Peter Dodds}.} \bibinfo{year}{2015}\natexlab{}.
\newblock \showarticletitle{Vaporous Marketing: Uncovering Pervasive Electronic
  Cigarette Advertisements on Twitter}.
\newblock \bibinfo{journal}{{\em arXiv preprint arXiv:1508.01843\/}}
  (\bibinfo{year}{2015}).
\newblock


\bibitem[\protect\citeauthoryear{Clark, Williams, Jones, Galbraith, Danforth,
  and Dodds}{Clark et~al\mbox{.}}{2016}]%
        {clark2016sifting}
\bibfield{author}{\bibinfo{person}{Eric Clark}, \bibinfo{person}{Jake
  Williams}, \bibinfo{person}{Chris Jones}, \bibinfo{person}{Richard
  Galbraith}, \bibinfo{person}{Christopher Danforth}, {and}
  \bibinfo{person}{Peter Dodds}.} \bibinfo{year}{2016}\natexlab{}.
\newblock \showarticletitle{Sifting robotic from organic text: a natural
  language approach for detecting automation on Twitter}.
\newblock \bibinfo{journal}{{\em Journal of Computational Science\/}}
  \bibinfo{volume}{16} (\bibinfo{year}{2016}), \bibinfo{pages}{1--7}.
\newblock


\bibitem[\protect\citeauthoryear{Cresci, Di~Pietro, Petrocchi, Spognardi, and
  Tesconi}{Cresci et~al\mbox{.}}{2017}]%
        {cresci2017paradigm}
\bibfield{author}{\bibinfo{person}{Stefano Cresci}, \bibinfo{person}{Roberto
  Di~Pietro}, \bibinfo{person}{Marinella Petrocchi}, \bibinfo{person}{Angelo
  Spognardi}, {and} \bibinfo{person}{Maurizio Tesconi}.}
  \bibinfo{year}{2017}\natexlab{}.
\newblock \showarticletitle{The paradigm-shift of social spambots: Evidence,
  theories, and tools for the arms race}. In \bibinfo{booktitle}{{\em
  Proceedings of the 26th International Conference on World Wide Web
  Companion}}. International World Wide Web Conferences Steering Committee,
  \bibinfo{pages}{963--972}.
\newblock


\bibitem[\protect\citeauthoryear{Davis, Varol, Ferrara, Flammini, and
  Menczer}{Davis et~al\mbox{.}}{2016}]%
        {davis2016botornot}
\bibfield{author}{\bibinfo{person}{Clayton~Allen Davis}, \bibinfo{person}{Onur
  Varol}, \bibinfo{person}{Emilio Ferrara}, \bibinfo{person}{Alessandro
  Flammini}, {and} \bibinfo{person}{Filippo Menczer}.}
  \bibinfo{year}{2016}\natexlab{}.
\newblock \showarticletitle{Botornot: A system to evaluate social bots}. In
  \bibinfo{booktitle}{{\em Proceedings of the 25th International Conference
  Companion on World Wide Web}}. International World Wide Web Conferences
  Steering Committee, \bibinfo{pages}{273--274}.
\newblock


\bibitem[\protect\citeauthoryear{Echeverr{\'\i}a and Zhou}{Echeverr{\'\i}a and
  Zhou}{2017}]%
        {zhou2017starwars}
\bibfield{author}{\bibinfo{person}{Juan Echeverr{\'\i}a} {and}
  \bibinfo{person}{Shi Zhou}.} \bibinfo{year}{2017}\natexlab{}.
\newblock \showarticletitle{The `Star Wars' botnet with >350k Twitter bots}.
\newblock \bibinfo{journal}{{\em arXiv preprint arXiv:1701.02405\/}}
  (\bibinfo{year}{2017}).
\newblock


\bibitem[\protect\citeauthoryear{Elyashar, Fire, Kagan, and Elovici}{Elyashar
  et~al\mbox{.}}{2013}]%
        {elyashar2013homing}
\bibfield{author}{\bibinfo{person}{Aviad Elyashar}, \bibinfo{person}{Michael
  Fire}, \bibinfo{person}{Dima Kagan}, {and} \bibinfo{person}{Yuval Elovici}.}
  \bibinfo{year}{2013}\natexlab{}.
\newblock \showarticletitle{Homing socialbots: intrusion on a specific
  organization's employee using Socialbots}. In \bibinfo{booktitle}{{\em Proc.
  IEEE/ACM Intl. Conf. on Advances in Social Networks Analysis and Mining}}.
  \bibinfo{pages}{1358--1365}.
\newblock


\bibitem[\protect\citeauthoryear{Ferrara}{Ferrara}{2015}]%
        {ferrara2015manipulation}
\bibfield{author}{\bibinfo{person}{Emilio Ferrara}.}
  \bibinfo{year}{2015}\natexlab{}.
\newblock \showarticletitle{Manipulation and abuse on social media}.
\newblock \bibinfo{journal}{{\em ACM SIGWEB Newsletter\/}}
  \bibinfo{number}{Spring} (\bibinfo{year}{2015}), \bibinfo{pages}{4}.
\newblock


\bibitem[\protect\citeauthoryear{Ferrara}{Ferrara}{2017}]%
        {ferrara2017disinformation}
\bibfield{author}{\bibinfo{person}{Emilio Ferrara}.}
  \bibinfo{year}{2017}\natexlab{}.
\newblock \showarticletitle{Disinformation and social bot operations in the run
  up to the 2017 French presidential election}.
\newblock \bibinfo{journal}{{\em First Monday\/}} \bibinfo{volume}{22},
  \bibinfo{number}{8} (\bibinfo{year}{2017}).
\newblock


\bibitem[\protect\citeauthoryear{Ferrara, Varol, Davis, Menczer, and
  Flammini}{Ferrara et~al\mbox{.}}{2016}]%
        {ferrara2016rise}
\bibfield{author}{\bibinfo{person}{Emilio Ferrara}, \bibinfo{person}{Onur
  Varol}, \bibinfo{person}{Clayton Davis}, \bibinfo{person}{Filippo Menczer},
  {and} \bibinfo{person}{Alessandro Flammini}.}
  \bibinfo{year}{2016}\natexlab{}.
\newblock \showarticletitle{The rise of social bots}.
\newblock \bibinfo{journal}{{\it Commun. ACM}} \bibinfo{volume}{59},
  \bibinfo{number}{7} (\bibinfo{year}{2016}), \bibinfo{pages}{96--104}.
\newblock


\bibitem[\protect\citeauthoryear{Forelle, Howard, Monroy-Hern{\'a}ndez, and
  Savage}{Forelle et~al\mbox{.}}{2015}]%
        {forelle2015political}
\bibfield{author}{\bibinfo{person}{Michelle~C Forelle},
  \bibinfo{person}{Philip~N Howard}, \bibinfo{person}{Andr{\'e}s
  Monroy-Hern{\'a}ndez}, {and} \bibinfo{person}{Saiph Savage}.}
  \bibinfo{year}{2015}\natexlab{}.
\newblock \showarticletitle{Political bots and the manipulation of public
  opinion in Venezuela}.
\newblock  (\bibinfo{year}{2015}).
\newblock


\bibitem[\protect\citeauthoryear{Gayo-Avello}{Gayo-Avello}{2017}]%
        {gayo2017social}
\bibfield{author}{\bibinfo{person}{Daniel Gayo-Avello}.}
  \bibinfo{year}{2017}\natexlab{}.
\newblock \showarticletitle{Social Media Won't Free Us}.
\newblock \bibinfo{journal}{{\em IEEE Internet Computing\/}}
  \bibinfo{volume}{21}, \bibinfo{number}{4} (\bibinfo{year}{2017}),
  \bibinfo{pages}{98--101}.
\newblock


\bibitem[\protect\citeauthoryear{Gers, Schmidhuber, and Cummins}{Gers
  et~al\mbox{.}}{1999}]%
        {gers1999learning}
\bibfield{author}{\bibinfo{person}{Felix~A Gers}, \bibinfo{person}{J{\"u}rgen
  Schmidhuber}, {and} \bibinfo{person}{Fred Cummins}.}
  \bibinfo{year}{1999}\natexlab{}.
\newblock \showarticletitle{Learning to forget: Continual prediction with
  LSTM}.
\newblock  (\bibinfo{year}{1999}).
\newblock


\bibitem[\protect\citeauthoryear{Goldberg}{Goldberg}{2016}]%
        {goldberg2016primer}
\bibfield{author}{\bibinfo{person}{Yoav Goldberg}.}
  \bibinfo{year}{2016}\natexlab{}.
\newblock \showarticletitle{A Primer on Neural Network Models for Natural
  Language Processing.}
\newblock \bibinfo{journal}{{\em J. Artif. Intell. Res.(JAIR)\/}}
  \bibinfo{volume}{57} (\bibinfo{year}{2016}), \bibinfo{pages}{345--420}.
\newblock


\bibitem[\protect\citeauthoryear{Gu, Zhang, and Lee}{Gu et~al\mbox{.}}{2008}]%
        {gu2008botsniffer}
\bibfield{author}{\bibinfo{person}{Guofei Gu}, \bibinfo{person}{Junjie Zhang},
  {and} \bibinfo{person}{Wenke Lee}.} \bibinfo{year}{2008}\natexlab{}.
\newblock \showarticletitle{BotSniffer: Detecting Botnet Command and Control
  Channels in Network Traffic.}. In \bibinfo{booktitle}{{\em NDSS}},
  Vol.~\bibinfo{volume}{8}. \bibinfo{pages}{1--18}.
\newblock


\bibitem[\protect\citeauthoryear{Gunning}{Gunning}{2017}]%
        {socialsim}
\bibfield{author}{\bibinfo{person}{David Gunning}.}
  \bibinfo{year}{2017}\natexlab{}.
\newblock \bibinfo{title}{DARPA: Explainable Artificial Intelligence (XAI)}.
\newblock   (\bibinfo{year}{2017}).
\newblock
\showURL{%
\url{https://www.darpa.mil/program/explainable-artificial-intelligence}}


\bibitem[\protect\citeauthoryear{Haustein, Bowman, Holmberg, Tsou, Sugimoto,
  and Larivi{\`e}re}{Haustein et~al\mbox{.}}{2016}]%
        {haustein2016tweets}
\bibfield{author}{\bibinfo{person}{Stefanie Haustein},
  \bibinfo{person}{Timothy~D Bowman}, \bibinfo{person}{Kim Holmberg},
  \bibinfo{person}{Andrew Tsou}, \bibinfo{person}{Cassidy~R Sugimoto}, {and}
  \bibinfo{person}{Vincent Larivi{\`e}re}.} \bibinfo{year}{2016}\natexlab{}.
\newblock \showarticletitle{Tweets as impact indicators: Examining the
  implications of automated ``bot'' accounts on Twitter}.
\newblock \bibinfo{journal}{{\em Journal of the Association for Information
  Science and Technology\/}} \bibinfo{volume}{67}, \bibinfo{number}{1}
  (\bibinfo{year}{2016}), \bibinfo{pages}{232--238}.
\newblock


\bibitem[\protect\citeauthoryear{He and Garcia}{He and Garcia}{2009}]%
        {he2009learning}
\bibfield{author}{\bibinfo{person}{Haibo He} {and} \bibinfo{person}{Edwardo~A
  Garcia}.} \bibinfo{year}{2009}\natexlab{}.
\newblock \showarticletitle{Learning from imbalanced data}.
\newblock \bibinfo{journal}{{\em IEEE Transactions on knowledge and data
  engineering\/}} \bibinfo{volume}{21}, \bibinfo{number}{9}
  (\bibinfo{year}{2009}), \bibinfo{pages}{1263--1284}.
\newblock


\bibitem[\protect\citeauthoryear{Hoang, Cohn, and Haffari}{Hoang
  et~al\mbox{.}}{2016}]%
        {hoang2016incorporating}
\bibfield{author}{\bibinfo{person}{Cong Duy~Vu Hoang}, \bibinfo{person}{Trevor
  Cohn}, {and} \bibinfo{person}{Gholamreza Haffari}.}
  \bibinfo{year}{2016}\natexlab{}.
\newblock \showarticletitle{Incorporating Side Information into Recurrent
  Neural Network Language Models.}. In \bibinfo{booktitle}{{\em HLT-NAACL}}.
  \bibinfo{pages}{1250--1255}.
\newblock


\bibitem[\protect\citeauthoryear{Hochreiter and Schmidhuber}{Hochreiter and
  Schmidhuber}{1997}]%
        {hochreiter1997long}
\bibfield{author}{\bibinfo{person}{Sepp Hochreiter} {and}
  \bibinfo{person}{J{\"u}rgen Schmidhuber}.} \bibinfo{year}{1997}\natexlab{}.
\newblock \showarticletitle{Long short-term memory}.
\newblock \bibinfo{journal}{{\em Neural computation\/}} \bibinfo{volume}{9},
  \bibinfo{number}{8} (\bibinfo{year}{1997}), \bibinfo{pages}{1735--1780}.
\newblock


\bibitem[\protect\citeauthoryear{Howard and Kollanyi}{Howard and
  Kollanyi}{2016}]%
        {howard2016bots}
\bibfield{author}{\bibinfo{person}{Philip~N Howard} {and}
  \bibinfo{person}{Bence Kollanyi}.} \bibinfo{year}{2016}\natexlab{}.
\newblock \showarticletitle{Bots,\# strongerin, and\# brexit: Computational
  propaganda during the uk-eu referendum}.
\newblock \bibinfo{journal}{{\em Browser Download This Paper\/}}
  (\bibinfo{year}{2016}).
\newblock


\bibitem[\protect\citeauthoryear{John}{John}{2017}]%
        {john2017survey}
\bibfield{author}{\bibinfo{person}{Vineet John}.}
  \bibinfo{year}{2017}\natexlab{}.
\newblock \showarticletitle{A Survey of Neural Network Techniques for Feature
  Extraction from Text}.
\newblock \bibinfo{journal}{{\em arXiv preprint arXiv:1704.08531\/}}
  (\bibinfo{year}{2017}).
\newblock


\bibitem[\protect\citeauthoryear{Jozefowicz, Zaremba, and Sutskever}{Jozefowicz
  et~al\mbox{.}}{2015}]%
        {jozefowicz2015empirical}
\bibfield{author}{\bibinfo{person}{Rafal Jozefowicz}, \bibinfo{person}{Wojciech
  Zaremba}, {and} \bibinfo{person}{Ilya Sutskever}.}
  \bibinfo{year}{2015}\natexlab{}.
\newblock \showarticletitle{An empirical exploration of recurrent network
  architectures}. In \bibinfo{booktitle}{{\em Proceedings of the 32nd
  International Conference on Machine Learning (ICML-15)}}.
  \bibinfo{pages}{2342--2350}.
\newblock


\bibitem[\protect\citeauthoryear{Karpathy, Johnson, and Fei-Fei}{Karpathy
  et~al\mbox{.}}{2015}]%
        {karpathy2015visualizing}
\bibfield{author}{\bibinfo{person}{Andrej Karpathy}, \bibinfo{person}{Justin
  Johnson}, {and} \bibinfo{person}{Li Fei-Fei}.}
  \bibinfo{year}{2015}\natexlab{}.
\newblock \showarticletitle{Visualizing and understanding recurrent networks}.
\newblock \bibinfo{journal}{{\em arXiv preprint arXiv:1506.02078\/}}
  (\bibinfo{year}{2015}).
\newblock


\bibitem[\protect\citeauthoryear{Krizhevsky, Sutskever, and Hinton}{Krizhevsky
  et~al\mbox{.}}{2012}]%
        {krizhevsky2012imagenet}
\bibfield{author}{\bibinfo{person}{Alex Krizhevsky}, \bibinfo{person}{Ilya
  Sutskever}, {and} \bibinfo{person}{Geoffrey~E Hinton}.}
  \bibinfo{year}{2012}\natexlab{}.
\newblock \showarticletitle{Imagenet classification with deep convolutional
  neural networks}. In \bibinfo{booktitle}{{\em Advances in neural information
  processing systems}}. \bibinfo{pages}{1097--1105}.
\newblock


\bibitem[\protect\citeauthoryear{LeCun, Bengio, and Hinton}{LeCun
  et~al\mbox{.}}{2015}]%
        {lecun2015deep}
\bibfield{author}{\bibinfo{person}{Yann LeCun}, \bibinfo{person}{Yoshua
  Bengio}, {and} \bibinfo{person}{Geoffrey Hinton}.}
  \bibinfo{year}{2015}\natexlab{}.
\newblock \showarticletitle{Deep learning}.
\newblock \bibinfo{journal}{{\em Nature\/}} \bibinfo{volume}{521},
  \bibinfo{number}{7553} (\bibinfo{year}{2015}), \bibinfo{pages}{436--444}.
\newblock


\bibitem[\protect\citeauthoryear{Lee, Eoff, and Caverlee}{Lee
  et~al\mbox{.}}{2011}]%
        {lee2011seven}
\bibfield{author}{\bibinfo{person}{Kyumin Lee}, \bibinfo{person}{Brian~David
  Eoff}, {and} \bibinfo{person}{James Caverlee}.}
  \bibinfo{year}{2011}\natexlab{}.
\newblock \showarticletitle{Seven Months with the Devils: A Long-Term Study of
  Content Polluters on Twitter.}. In \bibinfo{booktitle}{{\em Proc. 5th AAAI
  Intl. Conf. on Web and Social Media}}.
\newblock


\bibitem[\protect\citeauthoryear{Li, Chen, Hovy, and Jurafsky}{Li
  et~al\mbox{.}}{2015}]%
        {li2015visualizing}
\bibfield{author}{\bibinfo{person}{Jiwei Li}, \bibinfo{person}{Xinlei Chen},
  \bibinfo{person}{Eduard Hovy}, {and} \bibinfo{person}{Dan Jurafsky}.}
  \bibinfo{year}{2015}\natexlab{}.
\newblock \showarticletitle{Visualizing and understanding neural models in
  nlp}.
\newblock \bibinfo{journal}{{\em arXiv preprint arXiv:1506.01066\/}}
  (\bibinfo{year}{2015}).
\newblock


\bibitem[\protect\citeauthoryear{Loader and Mercea}{Loader and Mercea}{2011}]%
        {loader2011networking}
\bibfield{author}{\bibinfo{person}{Brian~D Loader} {and} \bibinfo{person}{Dan
  Mercea}.} \bibinfo{year}{2011}\natexlab{}.
\newblock \showarticletitle{Networking democracy? Social media innovations and
  participatory politics}.
\newblock \bibinfo{journal}{{\em Information, Communication \& Society\/}}
  \bibinfo{volume}{14}, \bibinfo{number}{6} (\bibinfo{year}{2011}),
  \bibinfo{pages}{757--769}.
\newblock


\bibitem[\protect\citeauthoryear{Metaxas and Mustafaraj}{Metaxas and
  Mustafaraj}{2012}]%
        {metaxas2012social}
\bibfield{author}{\bibinfo{person}{Panagiotis~T Metaxas} {and}
  \bibinfo{person}{Eni Mustafaraj}.} \bibinfo{year}{2012}\natexlab{}.
\newblock \showarticletitle{Social media and the elections}.
\newblock \bibinfo{journal}{{\em Science\/}} \bibinfo{volume}{338},
  \bibinfo{number}{6106} (\bibinfo{year}{2012}), \bibinfo{pages}{472--473}.
\newblock


\bibitem[\protect\citeauthoryear{Mikolov and Zweig}{Mikolov and Zweig}{2012}]%
        {mikolov2012context}
\bibfield{author}{\bibinfo{person}{Tomas Mikolov} {and}
  \bibinfo{person}{Geoffrey Zweig}.} \bibinfo{year}{2012}\natexlab{}.
\newblock \showarticletitle{Context dependent recurrent neural network language
  model.}
\newblock \bibinfo{journal}{{\em SLT\/}}  \bibinfo{volume}{12}
  (\bibinfo{year}{2012}), \bibinfo{pages}{234--239}.
\newblock


\bibitem[\protect\citeauthoryear{Miller, Dickinson, Deitrick, Hu, and
  Wang}{Miller et~al\mbox{.}}{2014}]%
        {miller2014twitter}
\bibfield{author}{\bibinfo{person}{Zachary Miller}, \bibinfo{person}{Brian
  Dickinson}, \bibinfo{person}{William Deitrick}, \bibinfo{person}{Wei Hu},
  {and} \bibinfo{person}{Alex~Hai Wang}.} \bibinfo{year}{2014}\natexlab{}.
\newblock \showarticletitle{Twitter spammer detection using data stream
  clustering}.
\newblock \bibinfo{journal}{{\em Information Sciences\/}}
  \bibinfo{volume}{260} (\bibinfo{year}{2014}), \bibinfo{pages}{64--73}.
\newblock


\bibitem[\protect\citeauthoryear{Mitter, Wagner, and Strohmaier}{Mitter
  et~al\mbox{.}}{2013}]%
        {mitter2014categorization}
\bibfield{author}{\bibinfo{person}{Silvia Mitter}, \bibinfo{person}{Claudia
  Wagner}, {and} \bibinfo{person}{Markus Strohmaier}.}
  \bibinfo{year}{2013}\natexlab{}.
\newblock \showarticletitle{A categorization scheme for socialbot attacks in
  online social networks}. In \bibinfo{booktitle}{{\em Proc. of the 3rd ACM Web
  Science Conference}}.
\newblock


\bibitem[\protect\citeauthoryear{Mnih, Kavukcuoglu, Silver, Graves, Antonoglou,
  Wierstra, and Riedmiller}{Mnih et~al\mbox{.}}{2013}]%
        {mnih2013playing}
\bibfield{author}{\bibinfo{person}{Volodymyr Mnih}, \bibinfo{person}{Koray
  Kavukcuoglu}, \bibinfo{person}{David Silver}, \bibinfo{person}{Alex Graves},
  \bibinfo{person}{Ioannis Antonoglou}, \bibinfo{person}{Daan Wierstra}, {and}
  \bibinfo{person}{Martin Riedmiller}.} \bibinfo{year}{2013}\natexlab{}.
\newblock \showarticletitle{Playing atari with deep reinforcement learning}.
\newblock \bibinfo{journal}{{\em arXiv preprint arXiv:1312.5602\/}}
  (\bibinfo{year}{2013}).
\newblock


\bibitem[\protect\citeauthoryear{Mnih, Kavukcuoglu, Silver, Rusu, Veness,
  Bellemare, Graves, Riedmiller, Fidjeland, Ostrovski, et~al\mbox{.}}{Mnih
  et~al\mbox{.}}{2015}]%
        {mnih2015human}
\bibfield{author}{\bibinfo{person}{Volodymyr Mnih}, \bibinfo{person}{Koray
  Kavukcuoglu}, \bibinfo{person}{David Silver}, \bibinfo{person}{Andrei~A
  Rusu}, \bibinfo{person}{Joel Veness}, \bibinfo{person}{Marc~G Bellemare},
  \bibinfo{person}{Alex Graves}, \bibinfo{person}{Martin Riedmiller},
  \bibinfo{person}{Andreas~K Fidjeland}, \bibinfo{person}{Georg Ostrovski},
  {et~al\mbox{.}}} \bibinfo{year}{2015}\natexlab{}.
\newblock \showarticletitle{Human-level control through deep reinforcement
  learning}.
\newblock \bibinfo{journal}{{\em Nature\/}} \bibinfo{volume}{518},
  \bibinfo{number}{7540} (\bibinfo{year}{2015}), \bibinfo{pages}{529--533}.
\newblock


\bibitem[\protect\citeauthoryear{M{\o}nsted, Sapie{\.z}y{\'n}ski, Ferrara, and
  Lehmann}{M{\o}nsted et~al\mbox{.}}{2017}]%
        {monsted2017evidence}
\bibfield{author}{\bibinfo{person}{Bjarke M{\o}nsted}, \bibinfo{person}{Piotr
  Sapie{\.z}y{\'n}ski}, \bibinfo{person}{Emilio Ferrara}, {and}
  \bibinfo{person}{Sune Lehmann}.} \bibinfo{year}{2017}\natexlab{}.
\newblock \showarticletitle{Evidence of Complex Contagion of Information in
  Social Media: An Experiment Using Twitter Bots}.
\newblock \bibinfo{journal}{{\em Plos One\/}} \bibinfo{volume}{12},
  \bibinfo{number}{9} (\bibinfo{year}{2017}).
\newblock


\bibitem[\protect\citeauthoryear{Pennington, Socher, and Manning}{Pennington
  et~al\mbox{.}}{2014}]%
        {pennington2014glove}
\bibfield{author}{\bibinfo{person}{Jeffrey Pennington},
  \bibinfo{person}{Richard Socher}, {and} \bibinfo{person}{Christopher~D.
  Manning}.} \bibinfo{year}{2014}\natexlab{}.
\newblock \showarticletitle{GloVe: Global Vectors for Word Representation}. In
  \bibinfo{booktitle}{{\em Empirical Methods in Natural Language Processing
  (EMNLP)}}. \bibinfo{pages}{1532--1543}.
\newblock
\showURL{%
\url{http://www.aclweb.org/anthology/D14-1162}}


\bibitem[\protect\citeauthoryear{Samek, Wiegand, and M{\"u}ller}{Samek
  et~al\mbox{.}}{2017}]%
        {samek2017explainable}
\bibfield{author}{\bibinfo{person}{Wojciech Samek}, \bibinfo{person}{Thomas
  Wiegand}, {and} \bibinfo{person}{Klaus-Robert M{\"u}ller}.}
  \bibinfo{year}{2017}\natexlab{}.
\newblock \showarticletitle{Explainable Artificial Intelligence: Understanding,
  Visualizing and Interpreting Deep Learning Models}.
\newblock \bibinfo{journal}{{\em arXiv preprint arXiv:1708.08296\/}}
  (\bibinfo{year}{2017}).
\newblock


\bibitem[\protect\citeauthoryear{Savage, Monroy-Hernandez, and
  H{\"o}llerer}{Savage et~al\mbox{.}}{2016}]%
        {savage2016botivist}
\bibfield{author}{\bibinfo{person}{Saiph Savage}, \bibinfo{person}{Andres
  Monroy-Hernandez}, {and} \bibinfo{person}{Tobias H{\"o}llerer}.}
  \bibinfo{year}{2016}\natexlab{}.
\newblock \showarticletitle{Botivist: Calling Volunteers to Action using Online
  Bots}. In \bibinfo{booktitle}{{\em Proceedings of the 19th ACM Conference on
  Computer-Supported Cooperative Work \& Social Computing}}. ACM,
  \bibinfo{pages}{813--822}.
\newblock


\bibitem[\protect\citeauthoryear{Shao, Ciampaglia, Varol, Flammini, and
  Menczer}{Shao et~al\mbox{.}}{2017}]%
        {shao2017spread}
\bibfield{author}{\bibinfo{person}{Chengcheng Shao},
  \bibinfo{person}{Giovanni~Luca Ciampaglia}, \bibinfo{person}{Onur Varol},
  \bibinfo{person}{Alessandro Flammini}, {and} \bibinfo{person}{Filippo
  Menczer}.} \bibinfo{year}{2017}\natexlab{}.
\newblock \showarticletitle{The spread of fake news by social bots}.
\newblock \bibinfo{journal}{{\em arXiv preprint arXiv:1707.07592\/}}
  (\bibinfo{year}{2017}).
\newblock


\bibitem[\protect\citeauthoryear{Silver, Huang, Maddison, Guez, Sifre, Van
  Den~Driessche, Schrittwieser, Antonoglou, Panneershelvam, Lanctot,
  et~al\mbox{.}}{Silver et~al\mbox{.}}{2016}]%
        {silver2016mastering}
\bibfield{author}{\bibinfo{person}{David Silver}, \bibinfo{person}{Aja Huang},
  \bibinfo{person}{Chris~J Maddison}, \bibinfo{person}{Arthur Guez},
  \bibinfo{person}{Laurent Sifre}, \bibinfo{person}{George Van Den~Driessche},
  \bibinfo{person}{Julian Schrittwieser}, \bibinfo{person}{Ioannis Antonoglou},
  \bibinfo{person}{Veda Panneershelvam}, \bibinfo{person}{Marc Lanctot},
  {et~al\mbox{.}}} \bibinfo{year}{2016}\natexlab{}.
\newblock \showarticletitle{Mastering the game of Go with deep neural networks
  and tree search}.
\newblock \bibinfo{journal}{{\em Nature\/}} \bibinfo{volume}{529},
  \bibinfo{number}{7587} (\bibinfo{year}{2016}), \bibinfo{pages}{484--489}.
\newblock


\bibitem[\protect\citeauthoryear{Stein, Chen, and Mangla}{Stein
  et~al\mbox{.}}{2011}]%
        {stein2011facebook}
\bibfield{author}{\bibinfo{person}{Tao Stein}, \bibinfo{person}{Erdong Chen},
  {and} \bibinfo{person}{Karan Mangla}.} \bibinfo{year}{2011}\natexlab{}.
\newblock \showarticletitle{Facebook immune system}. In
  \bibinfo{booktitle}{{\em Proc. of the 4th Workshop on Social Network
  Systems}}. ACM, \bibinfo{pages}{8}.
\newblock


\bibitem[\protect\citeauthoryear{Subrahmanian, Azaria, Durst, Kagan, Galstyan,
  Lerman, Zhu, Ferrara, Flammini, and Menczer}{Subrahmanian
  et~al\mbox{.}}{2016}]%
        {subrahmanian2016darpa}
\bibfield{author}{\bibinfo{person}{VS Subrahmanian}, \bibinfo{person}{Amos
  Azaria}, \bibinfo{person}{Skylar Durst}, \bibinfo{person}{Vadim Kagan},
  \bibinfo{person}{Aram Galstyan}, \bibinfo{person}{Kristina Lerman},
  \bibinfo{person}{Linhong Zhu}, \bibinfo{person}{Emilio Ferrara},
  \bibinfo{person}{Alessandro Flammini}, {and} \bibinfo{person}{Filippo
  Menczer}.} \bibinfo{year}{2016}\natexlab{}.
\newblock \showarticletitle{The DARPA Twitter bot challenge}.
\newblock \bibinfo{journal}{{\em Computer\/}} \bibinfo{volume}{49},
  \bibinfo{number}{6} (\bibinfo{year}{2016}), \bibinfo{pages}{38--46}.
\newblock


\bibitem[\protect\citeauthoryear{Szegedy, Liu, Jia, Sermanet, Reed, Anguelov,
  Erhan, Vanhoucke, and Rabinovich}{Szegedy et~al\mbox{.}}{2015}]%
        {szegedy2015going}
\bibfield{author}{\bibinfo{person}{Christian Szegedy}, \bibinfo{person}{Wei
  Liu}, \bibinfo{person}{Yangqing Jia}, \bibinfo{person}{Pierre Sermanet},
  \bibinfo{person}{Scott Reed}, \bibinfo{person}{Dragomir Anguelov},
  \bibinfo{person}{Dumitru Erhan}, \bibinfo{person}{Vincent Vanhoucke}, {and}
  \bibinfo{person}{Andrew Rabinovich}.} \bibinfo{year}{2015}\natexlab{}.
\newblock \showarticletitle{Going deeper with convolutions}. In
  \bibinfo{booktitle}{{\em Proceedings of the IEEE conference on computer
  vision and pattern recognition}}. \bibinfo{pages}{1--9}.
\newblock


\bibitem[\protect\citeauthoryear{Tomek}{Tomek}{1976}]%
        {tomek1976two}
\bibfield{author}{\bibinfo{person}{Ivan Tomek}.}
  \bibinfo{year}{1976}\natexlab{}.
\newblock \showarticletitle{Two modifications of CNN}.
\newblock \bibinfo{journal}{{\em IEEE Trans. Systems, Man and Cybernetics\/}}
  \bibinfo{volume}{6} (\bibinfo{year}{1976}), \bibinfo{pages}{769--772}.
\newblock


\bibitem[\protect\citeauthoryear{Wang, Konolige, Wilson, Wang, Zheng, and
  Zhao}{Wang et~al\mbox{.}}{2013a}]%
        {wang2013you}
\bibfield{author}{\bibinfo{person}{Gang Wang}, \bibinfo{person}{Tristan
  Konolige}, \bibinfo{person}{Christo Wilson}, \bibinfo{person}{Xiao Wang},
  \bibinfo{person}{Haitao Zheng}, {and} \bibinfo{person}{Ben~Y Zhao}.}
  \bibinfo{year}{2013}\natexlab{a}.
\newblock \showarticletitle{You are how you click: Clickstream analysis for
  sybil detection}. In \bibinfo{booktitle}{{\em Proc. USENIX Security}}.
  Citeseer, \bibinfo{pages}{1--15}.
\newblock


\bibitem[\protect\citeauthoryear{Wang, Mohanlal, Wilson, Wang, Metzger, Zheng,
  and Zhao}{Wang et~al\mbox{.}}{2013b}]%
        {wang2012social}
\bibfield{author}{\bibinfo{person}{Gang Wang}, \bibinfo{person}{Manish
  Mohanlal}, \bibinfo{person}{Christo Wilson}, \bibinfo{person}{Xiao Wang},
  \bibinfo{person}{Miriam Metzger}, \bibinfo{person}{Haitao Zheng}, {and}
  \bibinfo{person}{Ben~Y Zhao}.} \bibinfo{year}{2013}\natexlab{b}.
\newblock \showarticletitle{Social turing tests: Crowdsourcing sybil
  detection}. In \bibinfo{booktitle}{{\em Proc. of the 20th Network \&
  Distributed System Security Symposium (NDSS)}}.
\newblock


\bibitem[\protect\citeauthoryear{Wilson}{Wilson}{1972}]%
        {wilson1972asymptotic}
\bibfield{author}{\bibinfo{person}{Dennis~L Wilson}.}
  \bibinfo{year}{1972}\natexlab{}.
\newblock \showarticletitle{Asymptotic properties of nearest neighbor rules
  using edited data}.
\newblock \bibinfo{journal}{{\em IEEE Transactions on Systems, Man, and
  Cybernetics\/}} \bibinfo{volume}{2}, \bibinfo{number}{3}
  (\bibinfo{year}{1972}), \bibinfo{pages}{408--421}.
\newblock


\bibitem[\protect\citeauthoryear{Woolley}{Woolley}{2016}]%
        {woolley2016automating}
\bibfield{author}{\bibinfo{person}{Samuel~C Woolley}.}
  \bibinfo{year}{2016}\natexlab{}.
\newblock \showarticletitle{Automating power: Social bot interference in global
  politics}.
\newblock \bibinfo{journal}{{\em First Monday\/}} \bibinfo{volume}{21},
  \bibinfo{number}{4} (\bibinfo{year}{2016}).
\newblock


\bibitem[\protect\citeauthoryear{Yang, Wilson, Wang, Gao, Zhao, and Dai}{Yang
  et~al\mbox{.}}{2014}]%
        {yang2014uncovering}
\bibfield{author}{\bibinfo{person}{Zhi Yang}, \bibinfo{person}{Christo Wilson},
  \bibinfo{person}{Xiao Wang}, \bibinfo{person}{Tingting Gao},
  \bibinfo{person}{Ben~Y Zhao}, {and} \bibinfo{person}{Yafei Dai}.}
  \bibinfo{year}{2014}\natexlab{}.
\newblock \showarticletitle{Uncovering social network sybils in the wild}.
\newblock \bibinfo{journal}{{\em ACM Trans. Knowledge Discovery from Data\/}}
  \bibinfo{volume}{8}, \bibinfo{number}{1} (\bibinfo{year}{2014}),
  \bibinfo{pages}{2}.
\newblock


\end{thebibliography}

\end{document}